\documentclass[a4paper,fleqn]{cas-sc}
\usepackage[numbers,square,sort&compress]{natbib}
\usepackage{graphicx}
\usepackage{multicol,multirow}
\usepackage{amsmath,amssymb,amsthm}
\usepackage[utf8]{inputenc}
\usepackage[T1]{fontenc}
\usepackage[australian]{babel}
\usepackage{tabularx}
\usepackage{algorithmicx}
\usepackage{algpseudocode}
\usepackage{subfigure}
\usepackage{url}
\usepackage{cancel}
\usepackage{lineno}

\PassOptionsToPackage{unicode}{hyperref}
\PassOptionsToPackage{hyphens}{url}
\usepackage{setspace}
\usepackage{booktabs,array}
\usepackage{multirow}
\usepackage{etoolbox}
\usepackage{calc}
\usepackage{longtable}

\usepackage{color}
\usepackage{fancyvrb}

\DefineVerbatimEnvironment{Highlighting}{Verbatim}{commandchars=\\\{\}}
\usepackage{framed}
\definecolor{shadecolor}{RGB}{248,248,248}
\newenvironment{Shaded}{\begin{snugshade}}{\end{snugshade}}

\newcommand{\BuiltInTok}[1]{#1}

\newcommand{\CommentTok}[1]{\textcolor[rgb]{0.56,0.35,0.01}{\textit{#1}}}

\newcommand{\ControlFlowTok}[1]{\textcolor[rgb]{0.13,0.29,0.53}{\textbf{#1}}}

\newcommand{\DecValTok}[1]{\textcolor[rgb]{0.00,0.00,0.81}{#1}}

\newcommand{\ImportTok}[1]{#1}

\newcommand{\KeywordTok}[1]{\textcolor[rgb]{0.13,0.29,0.53}{\textbf{#1}}}
\newcommand{\NormalTok}[1]{#1}
\newcommand{\OperatorTok}[1]{\textcolor[rgb]{0.81,0.36,0.00}{\textbf{#1}}}

\newcommand{\SpecialCharTok}[1]{\textcolor[rgb]{0.00,0.00,0.00}{#1}}
\newcommand{\SpecialStringTok}[1]{\textcolor[rgb]{0.31,0.60,0.02}{#1}}
\newcommand{\StringTok}[1]{\textcolor[rgb]{0.31,0.60,0.02}{#1}}

\usepackage{xcolor}

\usepackage{verbatim}

\definecolor{blue2b}{rgb}{0,0.1,0.3}
\definecolor{blue2}{rgb}{0,0.2,0.7}
\definecolor{red2}{rgb}{0.6,0.1,0.0}
\definecolor{green2}{rgb}{0.1,0.4,0.0}
\definecolor{yel2}{rgb}{0.3,0.2,0.0}
\definecolor{purple2}{rgb}{0.5,0.0,0.5}
\definecolor{blue3}{rgb}{0.65,0.85,1.0}
\definecolor{red3}{rgb}{1.0,0.7,0.5}
\definecolor{green3}{rgb}{0.8,1.0,0.7}
\definecolor{yel3}{rgb}{1.0,1.0,0.7}
\definecolor{grey3}{rgb}{0.95,0.95,0.95}
\definecolor{gray3}{rgb}{0.95,0.95,0.95}

\definecolor{grey1}{rgb}{0.30,0.30,0.30}
\definecolor{grey2}{rgb}{0.20,0.20,0.20}
\definecolor{grey0}{rgb}{0,0,0}

\def\tsc#1{\csdef{#1}{\textsc{\lowercase{#1}}\xspace}}
\tsc{WGM}
\tsc{QE}
\tsc{EP}
\tsc{PMS}
\tsc{BEC}
\tsc{DE}

\makeatletter
\newtheoremstyle{definition}
{3ex}%
{3ex}%
{\upshape}%
{}%
{\bfseries}%
{.}%
{.5em}%
{\thmname{#1}\thmnumber{ #2}\thmnote{ (#3)}}
\makeatother

\theoremstyle{plain}

\theoremstyle{definition}

\ExplSyntaxOn
\keys_set:nn { stm / mktitle } { nologo }
\ExplSyntaxOff

\begin{document}
\let\WriteBookmarks\relax
\def\floatpagepagefraction{1}
\def\textpagefraction{.001}

\shorttitle{A Framework for Benchmarking Clustering Algorithms}
\title[mode = title]{A Framework for Benchmarking Clustering Algorithms}

\shortauthors{Gagolewski}

\author[1,2]{Marek Gagolewski}[orcid=0000-0003-0637-6028]
\ead{m.gagolewski@deakin.edu.au}
\ead[url]{https://www.gagolewski.com}

\address[1]{Warsaw University of Technology,
Faculty of Mathematics and Information Science,
ul. Koszykowa 75, 00-662 Warsaw, Poland}

\address[2]{Deakin University, Data to Intelligence Research Centre, School of IT, Geelong, VIC 3220, Australia}

\begin{abstract}
The evaluation of clustering algorithms can involve running them on a variety of benchmark problems, and comparing their outputs to the reference, ground-truth groupings provided by experts. Unfortunately, many research papers and graduate theses consider only a small number of datasets. Also, the fact that there can be many equally valid ways to cluster a given problem set is rarely taken into account. In order to overcome these limitations, we have developed a framework whose aim is to introduce a consistent methodology for testing clustering algorithms. Furthermore, we have aggregated, polished, and standardised many clustering benchmark dataset collections referred to across the machine learning and data mining literature, and included new datasets of different dimensionalities, sizes, and cluster types. An interactive datasets explorer, the documentation of the Python API, a description of the ways to interact with the framework from other programming languages such as R or MATLAB, and other details are all provided at \url{https://clustering-benchmarks.gagolewski.com}.
\end{abstract}


\maketitle

\ignorespaces\noindent\textit{
Please cite this paper as:
Gagolewski M., A framework for benchmarking clustering algorithms, {\normalfont SoftwareX} \textbf{20}, 101270, 2022, DOI:10.1016/j.softx.2022.101270.
This preprint includes some minor corrections.
}

\footnotesize

\begin{longtable}[]{@{}
  >{\raggedright\arraybackslash}p{(\columnwidth - 2\tabcolsep) * \real{0.3525}}
  >{\raggedright\arraybackslash}p{(\columnwidth - 2\tabcolsep) * \real{0.6475}}@{}}
\toprule
\begin{minipage}[b]{\linewidth}\raggedright
\textbf{Metadata}
\end{minipage} & \begin{minipage}[b]{\linewidth}\raggedright
\end{minipage} \\
\midrule
\endhead
Current code version & 1.1.2 \\
Permanent link to code repository &
\url{https://github.com/gagolews/clustering-benchmarks} \\
Legal code license & GNU AGPL v3 \\
Code versioning system used & git \\
Software code language used & Python \\
Compilation requirements, operating environments and dependencies & Python 3.7+
with \emph{numpy}, \emph{scipy}, \emph{pandas}, \emph{matplotlib}, \emph{scikit-learn}, and \emph{genieclust} \\
Link to developer documentation/manual &
\url{https://clustering-benchmarks.gagolewski.com} \\
Feature requests and bug tracker &
\url{https://github.com/gagolews/clustering-benchmarks/issues} \\
\bottomrule
\end{longtable}

\normalsize

\section{Introduction}
\label{sec:intro}

Cluster analysis
\cite{Hennig2015:whataretrueclusters,LuxburgETAL2012:clustscienceart,whitepaperclust} is a data
mining task where we discover \emph{semantically useful} dataset partitions in a
purely unsupervised manner. We know that there is no single ``best'' all-purpose
algorithm \cite{weightclust}, but some methods are better than others for
certain problem types. However, a lot is still yet to be done
\cite{xiongli2014:clustvalmeasures,objclustval,whitepaperclust} with regard to separating
promising approaches from the systematically disappointing ones.

One approach to clustering validation relies on using the so-called internal
measures, which are supposed to summarise the quality of partitions into a
single number
\cite{Milligan1985:psycho,Maulik2002:cvi_comp,ArbelaitzEtAl2013:extensive_CVI}.
In practice, they can only focus on a single property of a given split (e.g.,
set separability or compactness) and the partitions they promote might be far
from sound \cite{cvi}.

Another approach is to use the external validity measures
\cite{nca,psi,WagnerWagner2006:compclust,comphard} that quantify the similarity
between the generated clusterings and the reference (ground-truth) partitions
provided by experts.

Unfortunately, it is not rare for research papers and graduate theses to
consider only a small number of benchmark datasets. We regularly come across the
same 5--10 test problems from the UCI \cite{uci} database. This is obviously too
few to make any evaluation rigorous enough and thus may lead to overfitting \cite{overoptimistic,ullmanframework}. Some authors propose their own
datasets, but do not test their methods against other benchmark batteries. This
might give rise to biased conclusions, as there is a risk that only the problems
``easy'' for a method of interest were included. On the other hand, the
researchers who generously share their data (e.g.,
\cite{graves,fcps,ThrunUltsch2020:fcps,uci,kmsix}), unfortunately, might not
make the interaction with their batteries particularly smooth, as each of them
uses different file formats.

Furthermore, the existing repositories do not reflect the idea there might be
many equally valid/plausible/useful partitions of the same dataset; see
\cite{sdmc,LuxburgETAL2012:clustscienceart} for discussion.

On the other hand, some well-agreed-upon benchmark problems for a long time have
existed in other machine learning domains (classification and regression
datasets from the aforementioned UCI \cite{uci}; but also test functions for
testing global optimisation solvers, e.g.,
\cite{optimisation-benchmarks1,optimisation-benchmarks2}).

In order to overcome these gaps, the current project proposes a consistent
framework for benchmarking clustering algorithms. Its description is given in
the next section. Then, in Section~\ref{sec:examples}, we describe a Python API
(the \emph{clustering-benchmarks} package available at PyPI; see
\url{https://pypi.org/project/clustering-benchmarks/}) that makes the
interaction therewith relatively easy. Section~\ref{sec:conclusions} concludes
the paper and proposes a few ideas for the future evolution of this framework.

\section{Methodology}\label{sec:description}

We have compiled a quite large suite of example real and simulated
benchmark datasets.
For reproducibility, the releases of our suite are versioned: e.g.,
\url{https://github.com/gagolews/clustering-data-v1/releases/tag/v1.1.0} links
to a revision that has been published in September 2022
\cite{clustering_data_v1}.
Currently, there are nine batteries (collections), each of which features several
datasets of different origins, dimensionalities, size imbalancedness, and level of overlap; including, but not limited to,
\cite{kmsix,fcps,ThrunUltsch2020:fcps,ThrunStier2021:fcas,graves,uci,chameleon,bezdek_iris,hdbscanpkg,FrantiVirmajoki2006:ssets,SieranojaFranti2019:densitypeaks,psi,JainLaw2005:dilemma}\footnote{The original datasets were not equipped with alternative labellings nor with noise point markers; these were added by the current author.}; see the project's homepage for the detailed list.

Note that the datasets and the described software
are independent of each other.
Thanks to this, new datasets can easily be added in the future.
Also, the users are free to use their own collections
or access data from within other programming environments.
The current framework defines the suggested unified file format
which is detailed on the project's homepage.

\begin{figure}[t!]
\centering

\includegraphics{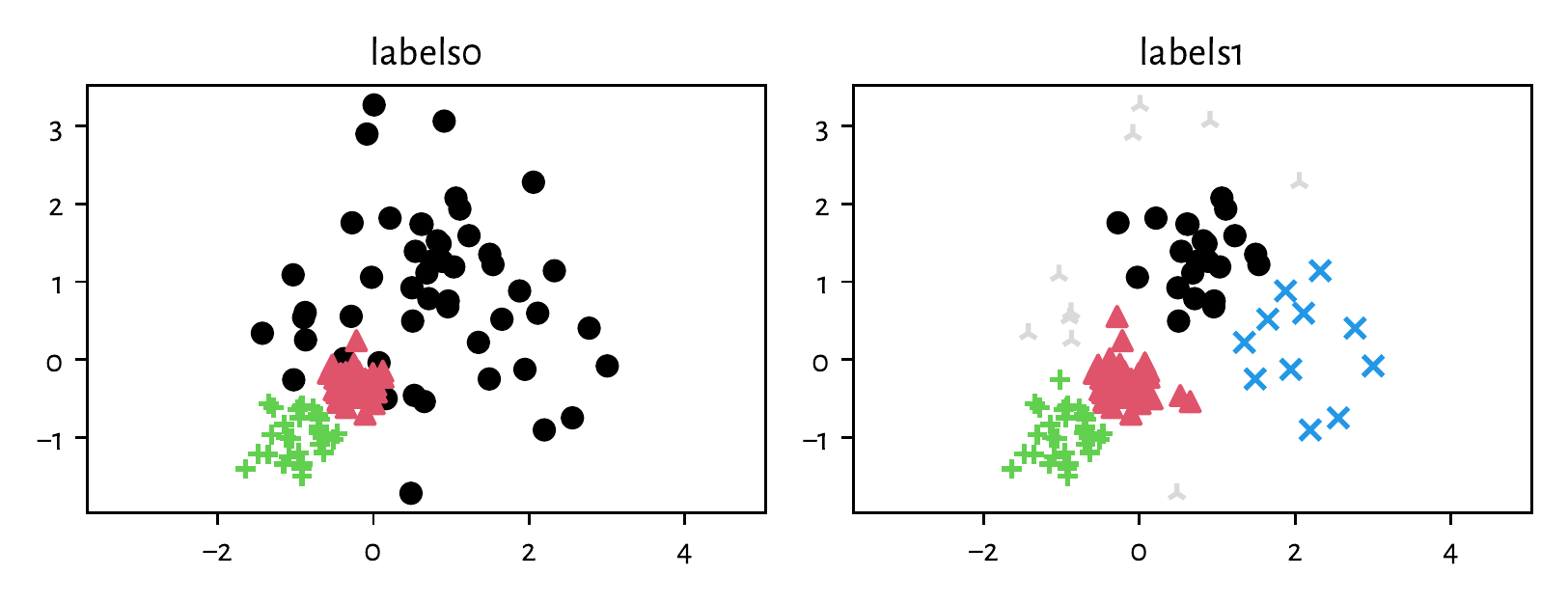}

\caption{
An example benchmark dataset (\textit{wut/x2}) and the two corresponding
reference partitions ($k=3$ and $k=4$; noise points marked in grey). \label{fig:example-dataset}
}
\end{figure}

\paragraph{Reference partitions.}
When referring to a particular benchmark problem, we use the convention
``\emph{battery/dataset}'', e.g, ``\emph{wut/x2}''. Let \(X\) be one of such
datasets that consists of \(n\) points in \(\mathbb{R}^d\).
Each dataset is equipped with a reference partition assigned by experts. Such a
grouping of the points into \(k \ge 2\) clusters is encoded using a label vector
\(\mathbf{y}\), where \(y_i\in\{1,\dots,k\}\) gives the cluster ID of the
\(i\)-th object. For instance, the left subfigure of
Figure~\ref{fig:example-dataset} depicts the ground-truth 3-clustering of
\emph{wut/x2} (which is based on the information about how this dataset has been
generated from a mixture of three Gaussian distributions).

\begin{figure}[t!]
\centering

\includegraphics{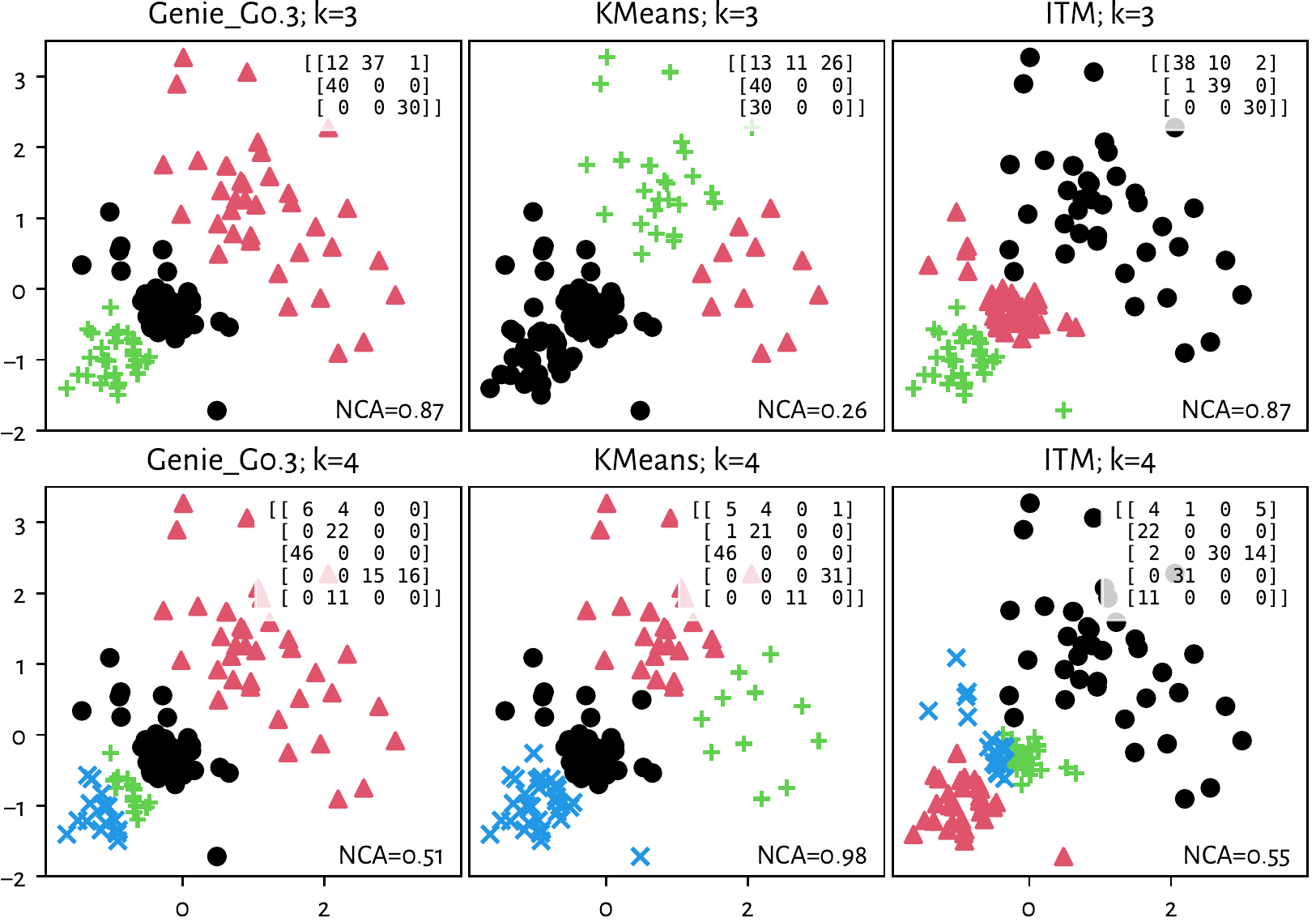}

\caption{
Clusterings of an example dataset (\textit{wut/x2})
discovered by Genie ($g=0.3$) \cite{genieins,genieclust}, k-means,
and ITM \cite{itm} ($k=3$ and $k=4$).
Confusion matrices and normalised clustering accuracies (NCA; Eq.~\eqref{eq:nca};
comparisons against the reference partitions depicted in Figure~\ref{fig:example-dataset})
are also reported. Note that the second ground-truth partition features
some noise points: hence, in the $k=4$ case, the first row of the confusion matrix is not taken into account.
\label{fig:example-results}
}
\end{figure}

\paragraph{Running the algorithm in question.}
Let us consider a clustering algorithm whose quality we would like to assess.
When we apply it on \(X\) to discover a new \(k\)-partition (in an unsupervised
manner, i.e., without revealing the true \(\mathbf{y}\)), we obtain a vector of
predicted labels encoding a new grouping, \(\hat{\mathbf{y}}\). For example, the
first row of scatterplots in Figure~\ref{fig:example-results} depicts the
3-partitions of \emph{wut/x2} discovered by three different methods.

\paragraph{Assessing partition similarity.}
Ideally, we would like to work with algorithms that yield partitions closely
matching the reference ones. This should be true on as wide a set of problems as
possible. Hence, we need to \emph{relate} the predicted labels to the reference
ones.

We can determine the confusion matrix \(\mathbf{C}\), where \(c_{i, j}\) denotes
the number of points in the \(i\)-th reference cluster that the algorithm
assigned to the \(j\)-th cluster. Even though such a matrix summarises all the
information required to judge the similarity between the two partitions, if we
wish to compare the quality of different algorithms, we would rather have it
aggregated in the form of a single number. As one of the many external cluster
validity indices (see, e.g., \cite{psi,WagnerWagner2006:compclust,comphard}), we
can use the normalised clustering accuracy \cite{nca} given by:

\begin{equation}\label{eq:nca}
\mathrm{NCA}(\mathbf{C})
=
\displaystyle\max_{\sigma:\{1,\dots,k\}\stackrel{\text{1--1}}{\to}\{1,\dots,k\}}
\frac{1}{k} \displaystyle\sum_{j=1}^{k} \frac{
    c_{\sigma(j),j}
    -
    \frac{1}{k} c_{\sigma(j),j}
}{
    c_{\sigma(j),\cdot}-\frac{1}{k} c_{\sigma(j),\cdot}
},
\end{equation}

\noindent which is the averaged percentage of correctly classified
points in each cluster \textit{above} the perfectly uniform label distribution.
As the actual cluster IDs do not matter (a partition is a \emph{set} of clusters
and sets are, by definition, unordered), the optimal matching between the
cluster labels is performed automatically by finding the best permutation
\(\sigma\) of the set \(\{1,\dots,k\}\).

\paragraph{There can be many valid partitions.}
What is more, it is in the very spirit of unsupervised learning that, in many
cases, there might be many equally valid ways to split a given dataset. An
algorithm should be rewarded for finding a partition that closely matches \emph{any} of the reference ones. This might require running the method multiple times
(unless it is a hierarchical one) to find the clusterings of different
cardinalities. Then, the generated outputs are evaluated against all the
available reference labellings and the \emph{maximal} similarity score is
reported.

\paragraph{Noise points.}
Also, to make the clustering problem more difficult, some datasets might feature
\emph{noise points} (e.g., outliers or irrelevant points in between the actual
clusters). They are specially marked in the ground-truth vectors: we assigned
them cluster IDs of 0; compare the right subfigure of
Figure~\ref{fig:example-dataset}, where they are coloured grey. A clustering
algorithm must never be informed about the location of such
``problematic'' points. Once the partition of the dataset is determined,
they are excluded from the computation of the external cluster validity measures.
In other words, it does not matter to which clusters the noise points are
allocated.

\section{The Python API}\label{sec:examples}

To facilitate the employment of the aforementioned framework, we have
implemented an open-source package for Python named
\emph{clustering-benchmarks}. It can be
installed from PyPI (\url{https://pypi.org/project/clustering-benchmarks/}),
e.g., via a call to \texttt{pip3\ install\ clustering-benchmarks}. Then, it can
be imported by calling:

\begin{Shaded}
\begin{Highlighting}[]
\ImportTok{import}\NormalTok{ clustbench  }\CommentTok{\# clustering{-}benchmarks}
\ImportTok{import}\NormalTok{ os.path, genieclust, sklearn.cluster  }\CommentTok{\# we will need these later}
\end{Highlighting}
\end{Shaded}

\paragraph{Fetching benchmark data.}
The example datasets repository \cite{clustering_data_v1}
(or any custom repository provided by the user) can be queried easily. Let us
assume that we store it in the following directory:

\begin{Shaded}
\begin{Highlighting}[]
\NormalTok{data\_path }\OperatorTok{=}\NormalTok{ os.path.join(}\StringTok{"\textasciitilde{}"}\NormalTok{, }\StringTok{"Projects"}\NormalTok{, }\StringTok{"clustering{-}data{-}v1"}\NormalTok{)  }\CommentTok{\# example}
\end{Highlighting}
\end{Shaded}

\noindent A particular dataset (here, for example: \emph{wut/x2}) can be accessed by calling:

\begin{Shaded}
\begin{Highlighting}[]
\NormalTok{battery, dataset }\OperatorTok{=} \StringTok{"wut"}\NormalTok{, }\StringTok{"x2"}
\NormalTok{b }\OperatorTok{=}\NormalTok{ clustbench.load\_dataset(battery, dataset, path}\OperatorTok{=}\NormalTok{data\_path)}
\end{Highlighting}
\end{Shaded}

\noindent The above call returns a named tuple, whose \texttt{data} field gives
the data matrix, \texttt{labels} gives the list of all ground-truth partitions
(encoded as label vectors), and \texttt{n\_clusters} gives the corresponding
numbers of subsets. For instance, here is a way in which we have generated
Figure~\ref{fig:example-dataset}.

\begin{Shaded}
\begin{Highlighting}[]
\ControlFlowTok{for}\NormalTok{ i }\KeywordTok{in} \BuiltInTok{range}\NormalTok{(}\BuiltInTok{len}\NormalTok{(b.labels)):}
\NormalTok{    plt.subplot(}\DecValTok{1}\NormalTok{, }\BuiltInTok{len}\NormalTok{(b.labels), i}\OperatorTok{+}\DecValTok{1}\NormalTok{)}
\NormalTok{    genieclust.plots.plot\_scatter(}
\NormalTok{        b.data, labels}\OperatorTok{=}\NormalTok{b.labels[i]}\OperatorTok{{-}}\DecValTok{1}\NormalTok{, axis}\OperatorTok{=}\StringTok{"equal"}\NormalTok{, title}\OperatorTok{=}\SpecialStringTok{f"labels}\SpecialCharTok{\{}\NormalTok{i}\SpecialCharTok{\}}\SpecialStringTok{"}\NormalTok{)}
\NormalTok{plt.show()}
\end{Highlighting}
\end{Shaded}

\paragraph{Fetching precomputed results.}
Suppose we would like to study some precomputed clustering results (see
\url{https://github.com/gagolews/clustering-results-v1}) which we store locally
in the following directory:

\begin{Shaded}
\begin{Highlighting}[]
\NormalTok{results\_path }\OperatorTok{=}\NormalTok{ os.path.join(}\StringTok{"\textasciitilde{}"}\NormalTok{, }\StringTok{"Projects"}\NormalTok{, }\StringTok{"clustering{-}results{-}v1"}\NormalTok{, }\StringTok{"original"}\NormalTok{)}
\end{Highlighting}
\end{Shaded}

\noindent The partitions can be fetched by calling:

\begin{Shaded}
\begin{Highlighting}[]
\NormalTok{res }\OperatorTok{=}\NormalTok{ clustbench.load\_results(}
    \StringTok{"Genie"}\NormalTok{, b.battery, b.dataset, b.n\_clusters, path}\OperatorTok{=}\NormalTok{results\_path}\NormalTok{)}
\BuiltInTok{print}\NormalTok{(}\BuiltInTok{list}\NormalTok{(res.keys()))}
\CommentTok{\#\# [\textquotesingle{}Genie\_G0.1\textquotesingle{}, \textquotesingle{}Genie\_G0.3\textquotesingle{}, \textquotesingle{}Genie\_G0.5\textquotesingle{}, \textquotesingle{}Genie\_G0.7\textquotesingle{}, \textquotesingle{}Genie\_G1.0\textquotesingle{}]}
\end{Highlighting}
\end{Shaded}

\noindent We thus have got access to data on the \emph{Genie}
\cite{genieins,genieclust} algorithm with different \texttt{gini\_threshold}
(\(g\)) parameter settings (\(g=1.0\) gives the single linkage method).

\paragraph{Computing external cluster validity measures.}
Here is a way to compute the external cluster validity measures:

\begin{Shaded}
\begin{Highlighting}[]
\BuiltInTok{round}\NormalTok{(clustbench.get\_score(b.labels, res[}\StringTok{"Genie\_G0.3"}\NormalTok{]), }\DecValTok{2}\NormalTok{)}
\CommentTok{\#\# 0.87}
\end{Highlighting}
\end{Shaded}

\noindent By default, the aforementioned normalised clustering accuracy (Eq.~\eqref{eq:nca})
is applied, but this might be changed to any other score by setting
the \texttt{metric} argument explicitly. As explained above, we compare the
predicted clusterings against all the reference partitions (ignoring the noise
points), and report the maximal score.

\paragraph{Applying clustering methods manually.}
We can use \texttt{clustbench.fit\_predict\_many} to generate all the partitions
required to compare ourselves against the reference labels. Let us test the
\emph{k}-means algorithm as implemented in the \emph{scikit-learn} package
\cite{sklearn}:

\begin{Shaded}
\begin{Highlighting}[]
\NormalTok{m }\OperatorTok{=}\NormalTok{ sklearn.cluster.KMeans()}
\NormalTok{res[}\StringTok{"KMeans"}\NormalTok{] }\OperatorTok{=}\NormalTok{ clustbench.fit\_predict\_many(m, b.data, b.n\_clusters)}
\BuiltInTok{round}\NormalTok{(clustbench.get\_score(b.labels, res[}\StringTok{"KMeans"}\NormalTok{]), }\DecValTok{2}\NormalTok{)}
\CommentTok{\#\# 0.98}
\end{Highlighting}
\end{Shaded}

\noindent We see that \emph{k}-means (which specialises in detecting symmetric
Gaussian-like blobs) performs better than \emph{Genie} on this particular
dataset; see Figure~\ref{fig:example-results} for an illustration (also
featuring the results generated by the ITM method \cite{itm}).

\bigskip The project's homepage and documentation discuss many more functions.

\section{Conclusion}\label{sec:conclusions}

The current project is designed to be extensible so that it can accommodate new
datasets and/or label vectors in the future --- so as to make the clustering
algorithm evaluation much more rigorous.
Any contributions are warmly
welcome; see \url{https://github.com/gagolews/clustering-benchmarks/issues} for
a feature request and bug tracker. In particular, we have implemented an
interactive standalone application that can be used for preparing our own
two-dimensional datasets (\emph{Colouriser}).

Future versions of the benchmark suite will include methods for generating
random samples of arbitrary sizes/cluster size distribution similar to a given
dataset (e.g., with more noise points).
Thanks to this, in the case of algorithms that feature many tunable parameters,
it will be possible to implement some means to separate validation
datasets (where we are allowed to learn the ``best'' settings;
see, e.g., \cite{ullmanframework} and the references therein)
from the testing ones (used in the final comparisons), which is a
quite standard approach in other machine learning domains.

Moreover, the framework can be extended to cover overlapping clusterings
as well as semi-supervised learning tasks,
where an algorithm knows about the right assignment of some of the input points
in advance.

\section*{Acknowledgements and Data Availability}

This research was supported by the Australian Research Council Discovery
Project ARC DP210100227.

Documentation and data are publicly available at
\url{https://clustering-benchmarks.gagolewski.com},
\url{https://github.com/gagolews/clustering-data-v1},
and
\url{https://github.com/gagolews/clustering-results-v1}.
A big thank-you to all the researchers who share their datasets with the clustering community.

\end{document}